\documentclass{article}
\usepackage{spconf}
\usepackage{xcolor}
\usepackage{subcaption}

\title{CoVariance Filters and Neural Networks \\ over Hilbert Spaces\thanks{\hspace{-.4cm} 
Work partially funded by the TU Delft AI Labs programme, the NWO OTP GraSPA proposal \#19497, the NWO VENI proposal 222.032. 
Email to: cbattiloro@hsph.harvard.edu}}

\name{Claudio~Battiloro$^1$, Andrea Cavallo$^2$ and Elvin~Isufi$^2$}
\address{$^1$Harvard University \quad $^2$Delft University of Technology}

\usepackage[T1]{fontenc}
\usepackage[utf8]{inputenc}
\usepackage{lmodern}
\usepackage{amsmath,amsthm,amssymb,mathtools,bm}
\usepackage[hidelinks]{hyperref}
\usepackage{microtype}
\usepackage{color}
\usepackage{lipsum}
\usepackage{caption}
\newtheorem{definition}{Definition}

\newtheorem{theorem}{Theorem}
\newtheorem{proposition}{Proposition}
\newtheorem{corollary}{Corollary}
\newtheorem{remark}{Remark}

\newcommand{\Hcal}{\mathcal{H}}
\newcommand{\ip}[2]{\left\langle #1,\,#2\right\rangle_{\Hcal}}
\newcommand{\norm}[1]{\left\lVert #1\right\rVert_{\Hcal}}
\newcommand{\R}{\mathbb{R}}

\begin{document}
\ninept
\maketitle

\begin{abstract}
Covariance Neural Networks (VNNs) perform graph convolutions on the empirical covariance matrix of signals defined over finite-dimensional Hilbert spaces, motivated by robustness and transferability properties. Yet, little is known about how these arguments extend to infinite-dimensional Hilbert spaces. In this work, we take a first step by introducing a novel convolutional learning framework for signals defined over infinite-dimensional Hilbert spaces, centered on the (empirical) covariance operator. We constructively define Hilbert coVariance Filters (HVFs) and design Hilbert coVariance  Networks (HVNs) as stacks of HVF filterbanks with nonlinear activations. We propose a principled discretization procedure, and we prove that empirical HVFs can recover the Functional PCA (FPCA) of the filtered signals. We then describe the versatility of our framework with examples ranging from multivariate real-valued functions to reproducing kernel Hilbert spaces. Finally, we validate HVNs on both synthetic and real-world time-series classification tasks, showing robust performance compared to MLP and FPCA-based classifiers.
\end{abstract}


\section{Introduction}
Covariance-based processing underpins several signal processing and machine learning frameworks, such as data whitening, exposing principal directions, and capturing feature interdependence. Its benefits have been shown in a wide plethora of domains, from brain neuroscience to finance \cite{Bessadok2022GNNNeuro, Cardoso2020FinanceGraphs, WangAste2022ICAIF, Wang2023MHAGNN, Liao2022BIBM}. The covariance matrix spectrum implicitly captures the structure of a dataset via the \textit{principal components}, and said structure can be exploited via the PCA transform. While PCA learning is broadly adopted~\cite{Jolliffe2002PCA}, its practical implementation suffers from the inherent stochastic perturbations in the empirical covariance matrix estimated from finite data, which gets amplified when in high-dimension/low-sample regimes or with tightly clustered eigenvalues \cite{Paul2007Spiked, BaikBenArousPeche2005}. To address this, coVariance Neural Networks (VNNs) \cite{Sihag2022coVarianceNN} perform graph convolutions using the empirical covariance matrix as graph shift operator, successfully merging the representational power of PCA and the stability/transferability properties of Graph Neural Networks (GNNs) \cite{Gama2020bStability}. However, VNNs operate only on finite-dimensional Hilbert spaces and little is known about how to generalize their benefits in cases where signals are defined over infinite-dimensional Hilbert spaces. This is a common setting in, for example, time-series processing \cite{parzen1959statistical} or quantum machine learning \cite{biamonte2017quantum}.

\noindent\textbf{Related Works.}  Covariance-based data representation has been extensively studied from a graph signal processing perspective. For example, the works in \cite{Marques2017StationaryGraphProcesses,Perraudin2017StationarySignalProcessing} formalized stationarity for graph signals and related the graph Fourier transform with the spectrum of the data covariances. This link has been also exploited in \cite{Rui2016DimensionalityReductionGSP,Shahid2016FastRobustPCAOnGraphs} for graph-based dimensionality reduction. More recently,  \cite{Sihag2022coVarianceNN} introduced VNNs, which use the empirical covariance matrix as a graph shift operator, and showed that VNNs are more expressive and stable than PCA. The transferability properties of VNNs have been studied in \cite{Sihag2023Transferability}. VNNs have been successfully applied to brain data in \cite{Sihag2024Explainable}. A variant of VNNs for streaming spatiotemporal signals has been proposed in \cite{cavallo2024spatiotemporal}, while VNNs with explicit bias mitigation have been proposed in \cite{cavallo2025fair}. The work in \cite{roy2025covariance} recently proposed to replace the empirical covariance matrix in VNNs with the empirical covariance density operators to improve user's control on the stability-discriminability trade-off of the network, and enhance robustness to noise. However, these works consider data only on finite-dimensional Hilbert spaces.


\noindent\textbf{Contribution.} We introduce a novel convolutional learning framework for random signals defined over general, possibly infinite-dimensional, Hilbert spaces, centered on the covariance operator of such space. We leverage the eigensystem of the covariance operator as a frequency domain to define a Hilbert coVariance Fourier Transform, and then we propose spectral Hilbert coVariance Filters (HVF) to manipulate the signal representation in such spectral domain.  We show how one can seamlessly develop Hilbert coVariance Networks (HVNs) as a layered architecture of HVFs and nonlinear activations in the same spirit as GNNs \cite{Gama2020bStability}. We propose a principled discretization procedure to make HVFs and HVNs implementable, and we prove that empirical HVFs can recover the Functional PCA (FPCA) \cite{shang2014surveyfpca} of the filtered signals. Finally, we numerically validate HVNs on both synthetic and real-world classification tasks of time-series, corroborating our findings and showing robustness benefits w.r.t. FPCA and covariance-unaware learning such as MLPs.


\section{Preliminaries and Motivation}
\label{sec:setting}

Consider a probability space $(\Omega,\mathcal{F},\mathbb{P})$ where $\Omega$ is the sample space (set of all possible outcomes), $\mathcal{F}$ is a sigma-algebra, i.e., a collection of measurable subsets of $\Omega$ treated as events, and $\mathbb{P}: \mathcal{F} \to [0,1]$ is a probability measure, i.e., a function assigning probabilities to the elements in $\mathcal{F}$. Consider also a separable Hilbert space $(\Hcal,\ip{\cdot}{\cdot})$ with inner product $\ip{\cdot}{\cdot}$  and norm  $\norm{\cdot}$. In this paper, we work with random variables $X:\Omega\to\Hcal$ that are square integrable, i.e.  they have finite expected squared norm $\mathbb{E}\,\norm{X}^2<\infty$. We denote the mean $\mu=\mathbb{E}[X]\in\Hcal$. We refer to the elements of $\Hcal$ as signals.


\noindent\textbf{Covariance operator.} \label{def:covop}
The  covariance operator $C:\Hcal\to\Hcal$
is the linear map that encodes all directional (mixed) second moments of $X$. The covariance operator acts on $v\in\Hcal$ as
\begin{equation}\label{eq:covop}
Cv \;=\; \mathbb{E}\Big[\,\underbrace{\ip{X-\mu}{v}}_{\text{scalar coefficient}}\,(X-\mu)\,\Big].
\end{equation}
Note that the operator $C$ is compact and trace-class \cite{royden1988real}, and hence there exist nonnegative eigenvalues $\{\lambda_\ell\}_{\ell\ge1}$ and orthonormal eigenfunctions $\{\phi_\ell\}_{\ell\ge1}$ such that 
\begin{equation}\label{eq:spectral-pop}
Cv \;=\; \sum_{\ell\ge1} \lambda_\ell\, \ip{v}{\phi_\ell}\,\phi_\ell.
\end{equation}



\noindent\textbf{Hilbert Covariance Fourier Transform (HVFT).}
A spectral representation of signals  is conventionally tied to a Fourier transform to analyze them. 
Given the eigen\,-system $\{(\lambda_\ell,\phi_\ell)\}_{\ell}$, we interpret the (ordered) eigenvalues as frequencies, and we define the Hilbert coVariance Fourier Transform (HVFT) of a signal $x\in\Hcal$ as the projection
\begin{equation}\label{eq:fvft}
\widetilde x[\ell] \;\triangleq\; \ip{x}{\phi_\ell}, \qquad \ell\geq1.
\end{equation}

\noindent Crucially, the HVFT of a signal $x \in \Hcal$ is equivalent to its Functional PCA (FPCA) transform \cite{shang2014surveyfpca} up to recentering. Although HVF/FPCA transforms are general means to obtain a different representation of signals, they also keep the usual variational characterization, i.e.,  taking the first $r$ eigenfunctions $\{\phi_\ell\}_\ell$ is equivalent to selecting the $r$ directions maximizing signal variance under orthonormality, or minimizing reconstruction error. In the finite-dimensional case ($\Hcal \cong \R^m$), FPCA and PCA (transforms) are equivalent.

\noindent\textbf{Problem Motivation.} As explained, the HVFT transform provides a principled spectral domain for (random) signals in Hilbert spaces.  Their spectral representations can be directly leveraged in processing/learning algorithms as information-rich features \cite{jolliffe1982note}. However, as we elaborate more in Sec. \ref{sec:discretization} and empirically show in Sec. \ref{sec:numerics}, practically implementing FPCA requires a finite number of discretized signals. This often leads to numerical instabilities and poor transferability performance, i.e., two finite batches of discretized samples coming from the same underlying Hilbert space or from similar Hilbert spaces could potentially results in significantly different spectral representations, making FPCA-only learning extremely difficult. For these reasons, there is a clear need for a proper notion of (convolutional) filtering \cite{vetterli2014foundations} and neural networks centered on the covariance operator, to successfully merge the representational power of FPCA with the stability/transferability properties of neural networks \cite{Gama2020bStability, Sihag2022coVarianceNN}.

\section{Hilbert Covariance Filters and Neural Networks}
In this section, we introduce spectral and spatial (polynomial) Hilbert coVariance Filters by building on the spectral domain induced by the covariance operator $C$. Then, we leverage these filters to design Hilbert coVariance Networks.

\begin{definition} [Spectral Hilbert coVariance Filter] Consider a function $h:[0,\norm{C}]\to\R$ bounded and Borel (e.g., continuous, monotone). A spectral Hilbert coVariance Filter (HVF) with frequency response $h$, denoted with $\mathbf{h}$, is defined as
\begin{equation}\label{eq:spectral-filter}
g = \mathbf{h}(C)\,x \;\triangleq\; \sum_{\ell=1}^{\infty} h(\lambda_\ell)\, \ip{x}{\phi_\ell}\,\phi_\ell + h(0)\,x_\perp,
\end{equation}
where $x \in \Hcal$ is the input and  $x_\perp$ is the orthogonal projection of $x$ onto $\ker(C)$.
\end{definition}
\noindent By direct computation, it is trivial to check that filtering is pointwise  in the HVFT domain, i.e., 
\begin{equation}\label{eq:fvf-response}
\widetilde{g}[\ell] \;=\; h(\lambda_\ell)\,\widetilde x[\ell], \qquad \ell\geq 1.
\end{equation}
This operation, where the frequency response $h(\lambda_\ell)$ amplifies or attenuates the different functional principal components, is in complete analogy with the convolutions/spectral filtering in several other domains, e.g., time \cite{vetterli2014foundations}, graphs \cite{isufi2024graph}, or manifolds \cite{battiloro2024tangenttsp}. By learning or designing $h(\lambda_\ell)$, we are able to put different weights on the different HVFT coefficients. As such, taking only the $r$ variance-maximizing directions, i.e., employing the FPCA transform as a dimensionality-reduction method, is then equivalent to an ideal high pass HVF. General spectral filters are maximally expressive, but they require explicit computation of eigenfunctions and eigenvalues.  For this reason, spatial (polynomial) filters are often preferred in practice, as they are (i) scalable and stable--they avoid eigendecomposition and are eigenvector-agnostic, and (ii) still expressive--polynomials are strong approximators.
\begin{definition}[Spatial Hilbert coVariance Filters] Given filter order $J\ge0$ and a set of real valued parameters $w_0,\ldots,w_J$, the corresponding spatial HVF is given by
\begin{equation}\label{eq:poly_filter}
\mathbf{h}(C) \;\triangleq\; \sum_{j=0}^{J} w_j\, C^{\,j}.
\end{equation}
\end{definition}
\noindent A spatial  HVF is entirely defined by the parameters $w_j$.  Indeed, using \eqref{eq:fvft}, it is trivial to show that the frequency response of a spatial HVF has the form $h(\lambda)=\sum_{k=0}^{J} w_k\,\lambda^k$. Intuitively, $C$ can be thought of as a shift operator (like the time delay \cite{vetterli2014foundations} or a graph shift \cite{isufi2024graph}).
Each power $C^{j}$ repeatedly spreads/smooths the signal through $j$-step covariance couplings (akin to $j$-hop diffusion), and the weights $w_j$ mix these scales. We can now define Hilbert coVariance  Networks using banks of HVFs and nonlinear activations.
\begin{definition} [Hilbert coVariance Networks] A Hilbert coVariance Network (HVN) is a layered architecture composed of banks of HVFs and nonlinear activations (operators). The $t-$th layer considers $F_t$ input signals $\left\{x_t^i\right\}_{i=1}^{F_t}$ and $F_{t+1}$ output signals $\left\{x_{t+1}^u\right\}_{u=1}^{F_{t+1}}$. Denoting the nonlinear activation with $\sigma:\Hcal\rightarrow \Hcal$ , the propagation rule reads as
\begin{equation}\label{eq:fvnn_layer}
x_{t+1}^u=\sigma\left(\sum_{i=1}^{F_t} \mathbf{h}(C)_t^{u, i} x_t^i\right), \quad u=1, \ldots, F_{t+1}.
\end{equation}
A HVN with input signals $\left\{x^i\right\}_{i=1}^{F_0}$ is then built as the stack of $T$ layers defined as in \eqref{eq:fvnn_layer}, where $x_0^i=x^i$. To globally represent the HVN, we collect all the filters in $\mathcal{W}=\left\{h_t^{u, i}\right\}_{t, u, i}$ and describe the HVN outputs as a mapping $\{x_T^u\}_{u=1}^{F_T}=\boldsymbol{\Psi}\left(\mathcal{W}, C,\left\{x^i\right\}_{i=1}^{F_0}\right)$ to enhance that it is parameterized by filters $\mathcal{H}$ and the covariance operator $C$. An HVN can be spectral or spatial, depending on the type of filters employed.
\end{definition}

\section{Discretization via Bounded Operators} \label{sec:discretization}

Hilbert coVariance Filters and Neural Networks operate on signals in a (possibly infinite-dimensional) Hilbert space whose distribution—and thus the covariance operator $C$—is unknown, so they are not directly implementable. We make them practical in two steps: (i) working with the empirical covariance operator using i.i.d., and (ii) discretizing the samples using bounded discretization operators to enable finite-dimensional processing and learning.

\noindent\textbf{Empirical covariance operator.}
Given i.i.d. samples $x_1,\ldots,x_n\in\Hcal$, let $\bar x = n^{-1}\sum_{i=1}^n x_i$. The empirical covariance operator $\hat C_n:\Hcal\to\Hcal$ is defined, for $v\in\Hcal$, by
\begin{equation}\label{eq:emp-cov}
\hat C_n v \;\triangleq\; \frac{1}{n} \sum_{i=1}^n \ip{x_i-\bar x}{v}\,(x_i-\bar x), \quad v\in\Hcal.
\end{equation}
By definition, $\hat C_n$ is bounded, self\,-adjoint, positive, and finite rank with $r:=\operatorname{rank}(\hat C_n)\le n-1$. Hence there exist eigenvalues $\hat\lambda_1\ge\cdots\ge\hat\lambda_r>0$ (with $r\le n-1$) and orthonormal eigenfunctions $\{\hat\phi_1,\ldots,\hat\phi_r\}$ such that 
\begin{equation}\label{eq:spectral-emp}
\hat C_n v \;=\; \sum_{\ell=1}^{r} \hat\lambda_\ell\, \ip{v}{\hat\phi_\ell}\,\hat\phi_\ell, \qquad v\in\Hcal.
\end{equation}
The $0$ eigenvalue may have infinite multiplicity with null spaces given by $\{v\in\Hcal:\ip{v}{\hat\phi_\ell}=0\ \forall\,\ell\le r\}$. As a valuable expressivity analysis of spatial HVFs, we now show that their empirical counterpart, i.e., replacing $C$ with $\hat C_n$ in \eqref{eq:poly_filter}, can provably recover the (empirical) FPCA transform of the samples. Let $\{\hat\lambda_1,\ldots,\hat\lambda_r\}$ be the positive eigenvalues of $\hat C_n$ (with multiplicity) and $\{\phi_\ell\}_{\ell=1}^r$ the corresponding orthonormal family. For a distinct eigenvalue value $\alpha$, let $I(\alpha)=\{\ell\le r:\hat\lambda_\ell=\alpha\}$ and define the orthogonal projector $P_{\alpha}:\Hcal\to\Hcal$ by
\begin{equation}\label{eq:projector}
P_{\alpha} x \;\triangleq\; \sum_{\ell\in I(\alpha)} \ip{x}{\hat \phi_\ell}\,\hat \phi_\ell, \qquad x\in\Hcal.
\end{equation}
\begin{theorem}\label{thm:exact-poly}
Let $\Lambda=\{\alpha_1,\ldots,\alpha_q\}$ (with $q\leq r$) be the set of \emph{distinct} positive eigenvalues of $\hat C_n$. For each $\alpha\in\Lambda$ there exists a spatial HVF $\mathbf{h}_{\alpha}$ as in \eqref{eq:poly_filter} such that
\begin{equation}\label{eq:poly-proj}
\mathbf{h}_{\alpha}(\hat C_n)\,x \;=\; P_{\alpha} x, \qquad x\in\Hcal.
\end{equation}
Moreover, one can choose $\deg( \mathbf{h}_{\alpha})\le q$.
\end{theorem}
\noindent \textbf{Proof.} See Appendix A.
\begin{corollary}\label{cor:fpca-score}
Fix $\alpha\in\Lambda$ and any orthonormal basis $\{\hat \phi_\ell\}_{\ell\in I(\alpha)}$ of its eigenspace. For any $x\in\Hcal$, there exists a polynomial HVF $\mathbf{h}_{\alpha}$ as in \eqref{eq:poly_filter} such that
\begin{equation}\label{eq:score-recovery}
\ip{\,\mathbf{h}_{\alpha}(\hat C_n)x\,}{\hat \phi_\ell} \,=\, \ip{x}{\hat \phi_\ell}, \qquad \forall\, \ell\in I(\alpha),
\end{equation}
i.e., recovering the $\alpha$-th (empirical) FPCA score of $x$.
\end{corollary}
\noindent\begin{definition}[Discretization Operator] Let $\{f_j\}_{j=1}^m$, with $f_j:\Hcal \rightarrow \R$ be bounded linear functionals on $\Hcal$. By the Riesz representation theorem, there exist $s_j\in\Hcal$ with $f_j(x)=\ip{x}{s_j}$, $x \in \Hcal$. The discretization operator $S_m: \Hcal \rightarrow \R^m$ acts on $x\in \Hcal$ as
\begin{equation}\label{eq:discr_op}
S_mx = [f_1(x),\ldots,f_m(x)]^\top = [\ip{x}{s_1},\ldots,\ip{x}{s_m}]^\top.
\end{equation}
The Hilbert adjoint $S_m^{*}:\mathbb{R}^m\to\Hcal$ satisfies $\ip{x}{S_m^{*}a}=\langle S_mx,a\rangle_{\mathbb{R}^m}$ and it acts on $a\in \R^m$ as
\begin{equation}
S_m^{*}a\;=\;\sum_{j=1}^m a_j\,s_j,\qquad a=[a_1,\ldots,a_m]^\top .
\end{equation}
\end{definition}
\noindent Given $x_1,\ldots,x_n\in\Hcal$ we denote $x_i^{(m)}:=S_mx_i\in\mathbb{R}^m$ and $\overline{x}^{(m)}=\tfrac1n\sum_i x_i^{(m)}=S_m\bar x$. Therefore, the empirical covariance matrix of the discretized signals is given by
\begin{equation}\label{eq:cov_mat}
\hat{\mathbf{C}}_n^{(m)}=\frac1n\sum_{i=1}^n \big(x_i^{(m)}-\overline{x}^{(m)}\big)\big(x_i^{(m)}-\overline{x}^{(m)}\big)^\top\in\mathbb{R}^{m\times m}.
\end{equation}
As a sanity check for the empirical covariance matrix, the following result shows that the discretized version of the empirical covariance operator is the empirical covariance matrix. In other words, discretizing the samples and then computing the empirical covariance matrix is equivalent to computing the empirical covariance operator and then discretizing it.
\begin{proposition}\label{prop:compression}
Let $S_m$, $\hat C_n$, and $\hat{\mathbf{C}}_n^{(m)}$ be a discretization operator as in \eqref{eq:discr_op}, the empirical covariance operator as in \eqref{eq:emp-cov}, and the empirical covariance matrix as in \eqref{eq:cov_mat}, respectively. The following holds:
\begin{equation}
\;\hat{\mathbf{C}}_n^{(m)}= S_m\,\hat C_n\,S_m^{*}.
\end{equation}
\end{proposition}
\noindent\textbf{Proof.} See Appendix A.

\noindent\textbf{Discrete HVFs, and HVNs.} Once signals are discretized and the empirical covariance matrix is computed, discrete HVFs and HVNs are simply given by \eqref{eq:poly_filter} and \eqref{eq:fvnn_layer} with the covariance operator replaced by the empirical covariance matrix and the signals replaced by their discretized version, i.e. $\{x^{(m),u}_T\}_{u=1}^{F_T}=\boldsymbol{\Psi}(\mathcal{W}, \hat{\mathbf{C}}_n^{(m)}, \{x^{(m)}_i\}_{i=1}^{n})$. Discretized HVNs are principled instances of coVariance Neural Network (VNN) \cite{Sihag2022coVarianceNN} and, as a consequence, of Graph (convolutional) Neural Networks (GNNs) \cite{Gama2020bStability}. To better enhance this similarity, we rewrite the layer in \eqref{eq:fvnn_layer} in  matrix form by introducing the matrices $\mathbf{X}^{(m)}_t=\{x_{t,i}^{(m)}\}_{i=1}^{F_t}\in \mathbb{R}^{m \times F_{t}}$, and $\mathbf{W}_{t,j} = \{w_{t,j}^{i,q}\}_{q=1,i=1}^{F_l,F_{t+1}}\in \mathbb{R}^{F_l \times F_{t+1}}$, as
\begin{equation}\label{eq:hvn_matrix}
    \mathbf{X}^{(m)}_{t+1} = \sigma\Bigg(\sum_{j = 0}^J\Big(\hat{\mathbf{C}}_n^{(m)}\Big)^j\mathbf{X}_t^{(m)}\mathbf{W}_{t,j}\Bigg) \; \in \mathbb{R}^{m \times F_{t+1}},
\end{equation}
where the filter weights $\{\mathbf{W}_{t,j}\}_{t,j}$ are learnable parameters, and $\mathbf{X}^{(m)}_0=\{x_i^{(m)}\}_{i=1}^{n}\in \mathbb{R}^{m \times n}$.

\noindent \textbf{Use-Cases for (Discrete) HVNs.} To showcase the versatility of our framework, in Appendix B, 
 we illustrate three practical $(\Hcal,S)$ pairs together with admissible nonlinear activations $\sigma:\Hcal\to\Hcal$. In particular, we describe how HVNs can be applied to functional data (e.g. time series, as we also show in Sec. \ref{sec:numerics}), infinite sequences, and data defined over Reproducing Kernel Hilbert Spaces.

\begin{figure*}[t]
    \centering
    \begin{subfigure}{0.32\textwidth}
        \centering
        \includegraphics[width=\linewidth]{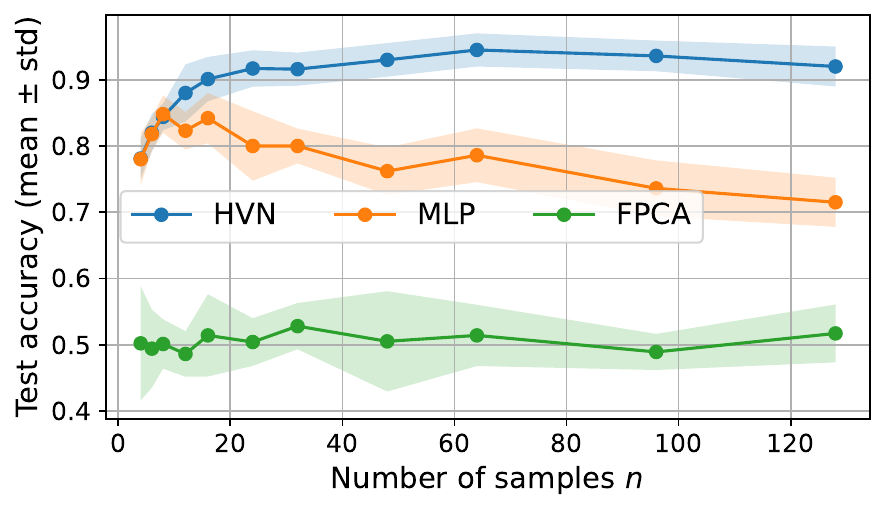}
        \caption{Test accuracy versus number of samples.}
        \label{Fig:acc_vs_n}
    \end{subfigure}\hfill
    \begin{subfigure}{0.32\textwidth}
        \centering 
        \includegraphics[width=\linewidth]{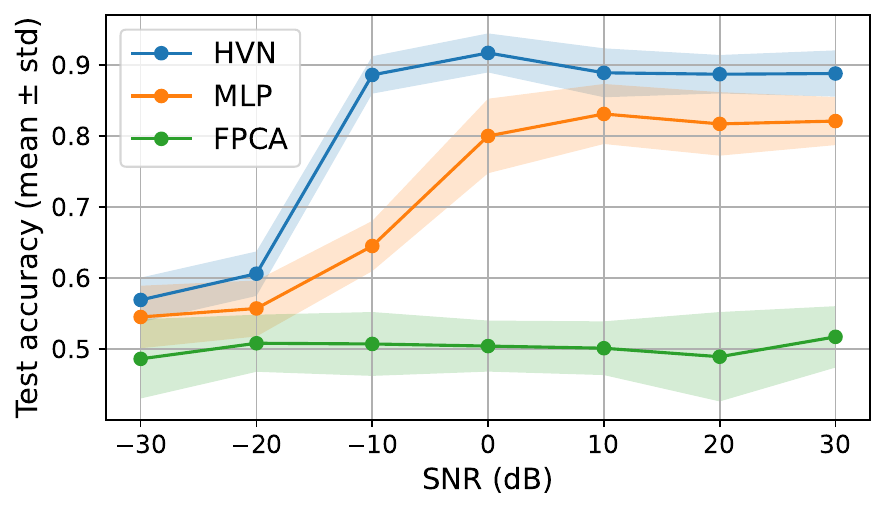}
        \caption{Test accuracy versus SNR (in dB).}
        \label{Fig:acc_vs_SNR}
    \end{subfigure}\hfill
    \begin{subfigure}{0.32\textwidth}
        \centering
        \includegraphics[width=\linewidth]{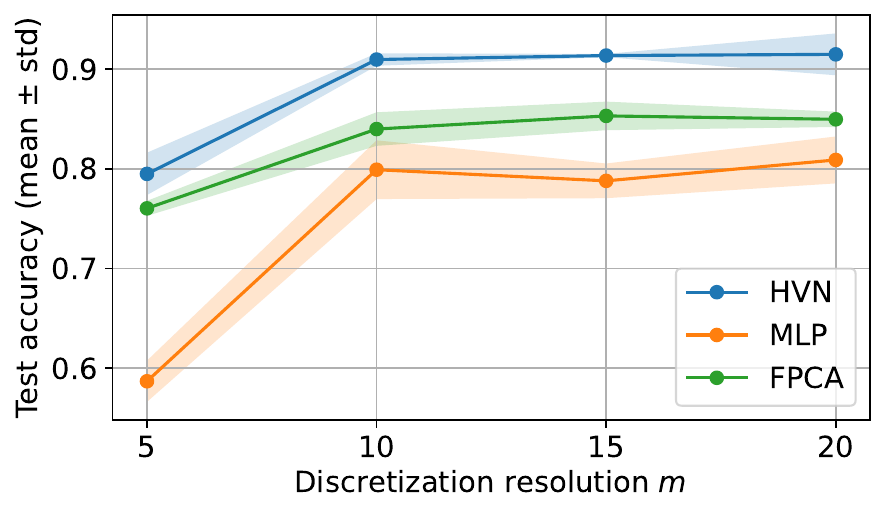}
        \caption{Test accuracy on ECG dataset.}
        \label{Fig:ecg}
    \end{subfigure}
\end{figure*}

\section{Numerical Results}\label{sec:numerics}
\vspace{-.2cm}
We validate HVNs on a synthetic and a real task. 
In the synthetic setting, we perform binary classification of multivariate time-series bags. 
The time-series are evaluated on the unit interval and have $d$ channels, i.e., they live in $\Hcal=L^2([0,1],\beta;\mathbb{R}^d)$, where $\beta$ is the Lebesgue measure. We observe them through the discretization operator $S_p$ as defined in Ex.~1 of Appendix B, 
i.e., we perform channelwise bin–averaging on a partition of $[0,1]$ into $p$ intervals, yielding $m=dp$ discrete components per sample. For a fixed class $y\in \{0,1\}$, the $i$-th sample $x_i$ is drawn from a zero–mean Gaussian process with matrix–valued kernel $K^{(y)}(t,s)=k_t(t,s)\Sigma_c^{(y)}\ \in\mathbb{R}^{d\times d}$, $0\le t,s\le 1$.
Equivalently, the covariance operator from \eqref{eq:covop} is given by
\begin{equation}\label{eq:Cy}
(C^{(y)}v)(t)=\int_0^1 K^{(y)}(t,s)v(s)ds,\qquad v\in\Hcal.
\end{equation}
HVNs leverage the whole covariance structure of the data. This means that, if the process is stationary, i.e., $k_t(t,s) = k_t(|t-s|)$, they are able to learn both from second order moments (the diagonal entries of the compressed covariance matrices) that are dependent only on a single channel at a certain time lag, and mixed second order moments (the off diagonal entries of the compressed covariance matrices) that are jointly dependent on all the channels at a certain time lag. For this reason, to structure a classification task whose discriminant elements can be captured by HVNs and not necessarily by models that only act component-wise, e.g., a linear model or an MLP,  we deliberately share the temporal kernel across classes and make the label depend only on cross–channel correlation, i.e.:
\begin{align}\label{eq:time-channel}
&k_t(t,s)^{(0)}= k_t(t,s)^{(1)} = \exp\Big(-\frac{(t-s)^2}{2\phi^2}\Big), \nonumber \\
&\textrm{and }\qquad\Sigma_c^{(0)}=I_d,\qquad
\Sigma_c^{(1)}=\big[\rho^{|i-j|}\big]_{i,j=1}^d,
\end{align}
where $\phi, \rho > 0$. 
In this task, a data point with class $y$ consists of a bag of $n$ i.i.d. samples drawn from
$x_1,\dots,x_n\ \stackrel{\text{i.i.d.}}{\sim}\ \mathsf{GP}\big(0,K^{(y)}\big),$
discretized as noisy $x_i^{(m)}:=S_m x_i + \epsilon \in\mathbb{R}^{m}$, where $\epsilon$ is AWGN, and centered by $\overline{x}^{(m)}=\tfrac1n\sum_{i=1}^n x_i^{(m)}$. We compare HVNs against two baselines acting on the same discretized bags: (i) MLP, obtained by simply setting $J = 1$ and $\hat{\mathbf{C}}_n^{(m)} = I_m$ in \eqref{eq:fvnn_layer}, and (ii) FPCA, consisting of computing the first $O$ discrete FPCA scores (i.e., the PCA scores of the discrete $\{x_i^{(m)}\}_n$ in the bag). For all the tested models, the output of the last layer (for HVN and MLP) or the scores (for PCA) are fed to an MLP classifier to get the class probabilities. 
We set $\phi=0.20$, and $\rho=0.7$, train with ADAM optimizer to minimize cross–entropy loss, we normalize the compressed covariance matrices for training stability (for HVN), and we use GELU nonlinearity (for HVN and MLP), mean pooling, and a polynomial degree (for HVN) $J=2$.  For both HVN and MLP, we use 2 layers, 32 output signals in each HVN layer,  and we adjust the MLP hidden dimensions to match the number of parameters of HVN. We fix $d=4$ and $p=32$, and we generate balanced train/test sets for different numbers of samples $n$ and signal to noise ratios $SNR_{dB}$, defined, given a bag, as $10\log_{10}\left((1/n)\sum_{i=1}^n\| x_i^{(m)}\|^2/\textrm{var}(\epsilon)\right)$. 
In Figures \ref{Fig:acc_vs_n}-\ref{Fig:acc_vs_SNR} we show the test accuracy vs $n$ (for fixed $SNR_{dB}=30$) and vs $SNR_{dB}$ (for fixed $n=24$), respectively. The MLP achieves mediocre accuracies (and is prone to overfitting) as it never mixes the $m$ components of the samples, accessing only first–order information, whereas the label is encoded in cross–feature covariance. The FPCA baseline gives random accuracy as it rotates each bag to its variance–maximizing basis, leading to coordinate drift across samples (sign flips, permutations,...), which a single classifier must reconcile. In contrast, HVN shows strong performance thanks to the transferable nature of HVFs and their ability to capture discriminative second–order structure consistently across bags. We further evaluate HVNs on the ECG5000 dataset from the UCR repository~\cite{ucr_dataset}, which contains 5000 ECG samples of length 140 and 5 classes denoting the heartbeat type.
We again compare HVNs, MLP, and FPCA with bin averaging and varying discretization resolution $m$. The hyperparameters are set as in the synthetic experiment. In this case, the goal is to classify individual time-series and not (possibly streaming) bags; thus, both for HVNN and FPCA, we compute the empirical covariance matrix on the entire training set and then use it to classify the individual time-series. In this way, both HVNN and FPCA have a more robust estimate of the covariance operator. Indeed, Figure~\ref{Fig:ecg} shows that HVN and FPCA outperform MLP across all values of $m$, indicating the importance of covariance information. Moreover, HVN outperforms both MLP and FPCA by a margin, especially for increasing $m$, corroborating its capability to robustly incorporate covariance information into its end-to-end learning dynamics.

\vspace{-.4cm}
\section{Conclusion}
\vspace{-.3cm}
We proposed a novel convolutional learning framework for signals in (possibly infinite-dimensional) Hilbert spaces, built around the (empirical) covariance operator. 
The generality of our framework suggests several future directions. We will study the convergence properties of discrete HVFs/HVNs to their infinite-dimensional counterparts as the discretization resolution becomes infinitely fine-grained. Using these results, we will analyze the transferability properties of HVNs. 

\bibliographystyle{IEEEbib}
\bibliography{refs}

\begin{thebibliography}{10}

\bibitem{Bessadok2022GNNNeuro}
Alaa Bessadok, Mohamed~Ali Mahjoub, and Islem Rekik,
\newblock ``Graph neural networks in network neuroscience,''
\newblock {\em IEEE Transactions on Pattern Analysis and Machine Intelligence},
  vol. 45, no. 5, pp. 5833--5848, 2022.

\bibitem{Cardoso2020FinanceGraphs}
Jos{\'e}~Vin{\'\i}cius de~Miranda~Cardoso, Jiaxi Ying, and Daniel~P{\'e}rez
  Palomar,
\newblock ``Algorithms for learning graphs in financial markets,''
\newblock {\em CoRR}, vol. abs/2012.15410, 2020.

\bibitem{WangAste2022ICAIF}
Yuanrong Wang and Tomaso Aste,
\newblock ``Network filtering of spatial-temporal gnn for multivariate
  time-series prediction,''
\newblock in {\em Proceedings of the 3rd ACM International Conference on AI in
  Finance (ICAIF '22)}, New York, NY, USA, 2022, pp. 463--470, ACM.

\bibitem{Wang2023MHAGNN}
Yan Wang, Xin Wang, Hongmei Yang, et~al.,
\newblock ``Mhagnn: A novel framework for wearable sensor-based human activity
  recognition combining multi-head attention and graph neural networks,''
\newblock {\em IEEE Transactions on Instrumentation and Measurement}, vol. 72,
  pp. 1--14, 2023.

\bibitem{Liao2022BIBM}
Tianzheng Liao, Jinjin Zhao, Yushi Liu, Kamen Ivanov, Jing Xiong, and Yan Yan,
\newblock ``Deep transfer learning with graph neural network for sensor-based
  human activity recognition,''
\newblock in {\em 2022 IEEE International Conference on Bioinformatics and
  Biomedicine (BIBM)}, 2022, pp. 2445--2452.

\bibitem{Jolliffe2002PCA}
Ian Jolliffe,
\newblock {\em Principal Component Analysis},
\newblock Springer, New York, 2002.

\bibitem{Paul2007Spiked}
Debashis Paul,
\newblock ``Asymptotics of sample eigenstructure for a large dimensional spiked
  covariance model,''
\newblock {\em Statistica Sinica}, vol. 17, no. 4, pp. 1617--1642, 2007.

\bibitem{BaikBenArousPeche2005}
Jinho Baik, G{\'e}rard~Ben Arous, and Sandrine P{\'e}ch{\'e},
\newblock ``Phase transition of the largest eigenvalue for nonnull complex
  sample covariance matrices,''
\newblock {\em The Annals of Probability}, vol. 33, no. 5, pp. 1643--1697,
  2005.

\bibitem{Sihag2022coVarianceNN}
Saurabh Sihag, Gonzalo Mateos, Corey~T. McMillan, and Alejandro Ribeiro,
\newblock ``covariance neural networks,''
\newblock in {\em Advances in Neural Information Processing Systems}, 2022,
  vol.~35, pp. 17003--17016.

\bibitem{Gama2020bStability}
Fernando Gama, Joan Bruna, and Alejandro Ribeiro,
\newblock ``Stability properties of graph neural networks,''
\newblock {\em IEEE Transactions on Signal Processing}, vol. 68, pp.
  5680--5695, 2020.

\bibitem{parzen1959statistical}
Emanuel Parzen,
\newblock ``Statistical inference on time series by hilbert space methods, i,''
\newblock Tech. {R}ep., 1959.

\bibitem{biamonte2017quantum}
Jacob Biamonte, Peter Wittek, Nicola Pancotti, Patrick Rebentrost, Nathan
  Wiebe, and Seth Lloyd,
\newblock ``Quantum machine learning,''
\newblock {\em Nature}, vol. 549, no. 7671, pp. 195--202, 2017.

\bibitem{Marques2017StationaryGraphProcesses}
Antonio~G. Marques, Santiago Segarra, Geert Leus, and Alejandro Ribeiro,
\newblock ``Stationary graph processes and spectral estimation,''
\newblock {\em IEEE Transactions on Signal Processing}, vol. 65, no. 22, pp.
  5911--5926, 2017.

\bibitem{Perraudin2017StationarySignalProcessing}
Nathana{\"e}l Perraudin and Pierre Vandergheynst,
\newblock ``Stationary signal processing on graphs,''
\newblock {\em IEEE Transactions on Signal Processing}, vol. 65, no. 13, pp.
  3462--3477, 2017.

\bibitem{Rui2016DimensionalityReductionGSP}
Li~Rui, Hemanth Nejat, and Ngai-Man Cheung,
\newblock ``Dimensionality reduction of brain imaging data using graph signal
  processing,''
\newblock in {\em 2016 IEEE International Conference on Image Processing
  (ICIP)}. IEEE, 2016, pp. 1329--1333.

\bibitem{Shahid2016FastRobustPCAOnGraphs}
Nika Shahid, Nathana{\"e}l Perraudin, Vassilis Kalofolias, Gilles Puy, and
  Pierre Vandergheynst,
\newblock ``Fast robust pca on graphs,''
\newblock {\em IEEE Journal of Selected Topics in Signal Processing}, vol. 10,
  no. 4, pp. 740--756, 2016.

\bibitem{Sihag2023Transferability}
Saurabh Sihag, Gonzalo Mateos, Corey~T. McMillan, and Alejandro Ribeiro,
\newblock ``Transferability of covariance neural networks and application to
  interpretable brain age prediction using anatomical features,''
\newblock {\em arXiv preprint arXiv:2305.01807}, 2023.

\bibitem{Sihag2024Explainable}
Saurabh Sihag, Gonzalo Mateos, Corey McMillan, and Alejandro Ribeiro,
\newblock ``Explainable brain age prediction using covariance neural
  networks,''
\newblock in {\em Advances in Neural Information Processing Systems (NeurIPS)},
  2024, vol.~36.

\bibitem{cavallo2024spatiotemporal}
Andrea Cavallo, Mohammad Sabbaqi, and Elvin Isufi,
\newblock ``Spatiotemporal covariance neural networks,''
\newblock in {\em Joint European Conference on Machine Learning and Knowledge
  Discovery in Databases}. Springer, 2024, pp. 18--34.

\bibitem{cavallo2025fair}
Andrea Cavallo, Madeline Navarro, Santiago Segarra, and Elvin Isufi,
\newblock ``Fair covariance neural networks,''
\newblock in {\em ICASSP 2025-2025 IEEE International Conference on Acoustics,
  Speech and Signal Processing (ICASSP)}. IEEE, 2025, pp. 1--5.

\bibitem{roy2025covariance}
Om~Roy, Yashar Moshfeghi, and Keith Smith,
\newblock ``Covariance density neural networks,''
\newblock {\em arXiv preprint arXiv:2505.11139}, 2025.

\bibitem{shang2014surveyfpca}
Han~Lin Shang,
\newblock ``A survey of functional principal component analysis,''
\newblock {\em AStA Advances in Statistical Analysis}, vol. 98, no. 2, pp.
  121--142, 2014.

\bibitem{royden1988real}
Halsey~Lawrence Royden and Patrick Fitzpatrick,
\newblock {\em Real analysis}, vol.~32,
\newblock Macmillan New York, 1988.

\bibitem{jolliffe1982note}
Ian~T Jolliffe,
\newblock ``A note on the use of principal components in regression,''
\newblock {\em Journal of the Royal Statistical Society Series C: Applied
  Statistics}, vol. 31, no. 3, pp. 300--303, 1982.

\bibitem{vetterli2014foundations}
Martin Vetterli, Jelena Kova{\v{c}}evi{\'c}, and Vivek~K Goyal,
\newblock {\em Foundations of signal processing},
\newblock Cambridge University Press, 2014.

\bibitem{isufi2024graph}
Elvin Isufi, Fernando Gama, David~I Shuman, and Santiago Segarra,
\newblock ``Graph filters for signal processing and machine learning on
  graphs,''
\newblock {\em IEEE Transactions on Signal Processing}, vol. 72, pp.
  4745--4781, 2024.

\bibitem{battiloro2024tangenttsp}
Claudio Battiloro, Zhiyang Wang, Hans Riess, Paolo Di~Lorenzo, and Alejandro
  Ribeiro,
\newblock ``Tangent bundle convolutional learning: from manifolds to cellular
  sheaves and back,''
\newblock {\em IEEE Transactions on Signal Processing}, vol. 72, pp.
  1892--1909, 2024.

\bibitem{ucr_dataset}
Hoang~Anh Dau, Anthony Bagnall, Kaveh Kamgar, Chin-Chia~Michael Yeh, Yan Zhu,
  Shaghayegh Gharghabi, Chotirat~Ann Ratanamahatana, and Eamonn Keogh,
\newblock ``The ucr time series archive,''
\newblock {\em IEEE/CAA Journal of Automatica Sinica}, vol. 6, no. 6, pp.
  1293--1305, November 2019.

\bibitem{parada2025convolutionalrkhs}
Alejandro Parada-Mayorga, Leopoldo Agorio, Alejandro Ribeiro, and Juan
  Bazerque,
\newblock ``Convolutional filtering with rkhs algebras,''
\newblock {\em IEEE Transactions on Signal Processing}, 2025.

\end{thebibliography}
\appendix
\section{Proofs}
\noindent\textbf{Proof of Theorem 1.} Define the scaled Lagrange basis on $\Lambda \cup \{0\}$ as
\begin{equation}\label{eq:lagrange}
L_{\alpha}(t)\;=\; \frac{t}{\alpha}\prod_{\beta\in\Lambda,\,\beta\ne\alpha} \cdot\frac{t-\beta}{\alpha-\beta} ,\qquad t\in\R.
\end{equation}
Then $L_{\alpha}(\alpha)=1$, $L_{\alpha}(\beta)=0$, and $L_{\alpha}(0)=0$ for all $\beta\in\Lambda\setminus\{\alpha\}$, and $L_\alpha$ is a polynomial of degree $\le q$. Its "sum-of-powers" form is given by
\begin{equation}\label{eq:lagrance_in_poly}
L_{\alpha}(t) \;=\; \sum_{k=0}^{q} w^{(\alpha)}_{k}\, t^{k},
\end{equation}
where the coefficients $w^{(\alpha)}_{k}$ are uniquely determined by the Vandermonde system
\begin{equation}\frac{1}{\alpha}
\begin{bmatrix}
1 & \alpha_1 & \cdots & \alpha_1^{\,q-1}\\
1 & \alpha_2 & \cdots & \alpha_2^{\,q-1}\\
\vdots & \vdots & \ddots & \vdots\\
1 & \alpha_q & \cdots & \alpha_q^{\,q-1}
\end{bmatrix}
\begin{bmatrix}w^{(\alpha)}_{0}\\ w^{(\alpha)}_{1}\\ \vdots\\ w^{(\alpha)}_{q-1}\end{bmatrix}
\;=\;
\begin{bmatrix}0\\ \cdots\\ 1\\ \cdots\\ 0\end{bmatrix}\!,
\end{equation}
where the $1$ on the RHS is in the row corresponding to $\alpha$.  Injecting \eqref{eq:lagrance_in_poly} in \eqref{eq:poly_filter}, for any $x\in\Hcal$, it then holds
\begin{align}\label{eq:apply-poly}
\mathbf{h}_{\alpha}(\hat C_n) x 
&= \sum_{\ell=1}^{r} h_{\alpha}(\hat\lambda_\ell)\, \ip{x}{\hat\phi_\ell}\,\hat\phi_\ell 
\;+\; h_{\alpha}(0)\,x_\perp\nonumber \\
&= \sum_{\ell\in I(\alpha)} \ip{x}{\hat\phi_\ell}\,\hat\phi_\ell 
= P_{\alpha} x.
\end{align}
\noindent\textbf{Proof of Proposition 1.}
Fix $a,b\in\R^m$. Using $x_i^{(m)}=S_mx_i$ and $\overline{x}^{(m)}=S_m\bar x$,
\begin{align*}
\langle a,\,\hat{\mathbf{C}}_n^{(m)} b\rangle_{\R^m}
&= \frac1n\sum_{i=1}^n \big\langle a,\; S_m(x_i-\bar x)\big\rangle_{\R^m}
                         \big\langle S_m(x_i-\bar x),\; b\big\rangle_{\R^m} \\
&\overset{(1)}{=} \frac1n\sum_{i=1}^n \ip{x_i-\bar x}{S_m^* a}\;
                              \ip{x_i-\bar x}{S_m^* b} \\
&\overset{(2)}{=} \left\langle \frac1n\sum_{i=1}^n
              \ip{x_i-\bar x}{S_m^* b}\,(x_i-\bar x)\;,\; S_m^* a \right\rangle_{\Hcal} \\
&\overset{(3)}{=} \ip{S_m^* a}{\hat C_n (S_m^* b)} \overset{(4)}{=} \big\langle a,\; S_m\,\hat C_n\,S_m^* b \big\rangle_{\R^m},
\end{align*}
where (1) uses the defining property of the adjoint, (2) uses linearity of the inner product, (3) applies the definition of $\hat C_n$ applied to the vector $S_m^* b$, and (4) applies the adjoint identity again with $u=\hat C_n S_m^* b$. Since this holds for all $a,b\in\R^m$, the bilinear forms induced by $\hat{\mathbf{C}}_n^{(m)}$ and $S_m\hat C_n S_m^*$ agree.

\section{Use-Cases for HVNs}\label{app:examples}

\noindent\textbf{[Ex. 1] Square-integrable Multivariate Real-valued Functions.}
Let $\Hcal=L^2(\R,\beta;\R^{d})$ with $\beta$ the Lebesgue measure. Let us denote with $\{B_j\}_{j=1}^{p}$ a collection of bins, i.e., a measurable partition with $\beta(B_j)>0$.
Set $m:=pd$ and fix any bijection $\iota:\{1,\ldots,p\}\times\{1,\ldots,d\}\to\{1,\ldots,m\}$ to flatten $(j,c)$-pairs into a single index.
Define $S_m:\Hcal\!\to\!\R^{m}$ to perform channelwise bin-averaging by
\begin{align}
(S_mv)_{\iota(j,c)} \;=\;& \frac{1}{\sqrt{\beta(B_j)}}\int_{B_j} v_c(t)\,d\beta(t), 
\; \nonumber\\
&j=1,\ldots,p, c=1,\ldots,d
\end{align}
so $S_m^* a=\sum_{j=1}^{p}\sum_{c=1}^{d} a_{\iota(j,c)}\,s_{(j,c)}$ with
\begin{equation}
s_{(j,c)}(t)\;=\;\beta(B_j)^{-1/2}\,\mathbf{1}_{B_j}(t)\,e_c \;\in\; L^2(\R,\beta;\R^{d}),
\end{equation}
where $\{e_c\}_{c=1}^d$ is the canonical basis of $\R^d$.
Typical nonlinearities are pointwise Lipschitz transforms
\begin{equation}
(\sigma v)(t) \;=\; \sigma_\ast\!\big(v(t)\big),
\end{equation}
with $\sigma_\ast:\R^d\!\to\!\R^d$ Lipschitz, e.g., componentwise ReLU. 

\noindent\textbf{Remark.} We constructively employed bin-averaging as a special intuitive case of a more general discretization approach. In particular, given any fixed orthonormal basis $\{\psi_j\}_{j\ge1}$ of $\Hcal$--in the case of bin averaging, the orthonormal basis implicitly employed is the set of normalized bin indicators per channel, $S_m:\Hcal\!\to\!\R^m$ can always be defined as "taking the first $m$ coefficients in that basis", i.e.,
\begin{equation}
(S_m x)_j \;=\; \ip{x}{\psi_j},\qquad j=1,\ldots,m,
\end{equation}
so $S_m^* a \;=\; \sum_{j=1}^m a_j\,\psi_j$. In other words,
$S_m^*S_m$ is the orthogonal projector onto $\mathrm{span}\{\psi_1,\ldots,\psi_m\}$.

\noindent\textbf{[Ex. 2] Square-summable Real-valued Sequences}
Let $\Hcal=\ell^2(\mathbb{N})$. $S_m:\ell^2\!\to\!\R^m$ can be defined as the canonical projection as
\begin{equation}
(S_mv)_j \;=\; v_j,\qquad j=1,\dots,m,
\end{equation}
so $S_m^* a=\sum_{j=1}^m a_j e_j$ and $S_m^*S_m=P_m$ is again the orthogonal projector onto the first $m$ coordinates. Typical nonlinearities are pointwise Lipschitz maps
\begin{equation}
(\sigma v)_k \;=\; \sigma_\ast(v_k).
\end{equation}

\noindent\textbf{[Ex. 3] Reproducing Kernel Hilbert Spaces.}
Let $\Hcal=\mathcal{H}_K$ be the reproducing kernel Hilbert space on a set $\mathcal{T}$ with kernel $K:\mathcal{T}\times\mathcal{T}\to\R$.
For sampling locations $\{t_j\}_{j=1}^m\subset\mathcal{T}$, point evaluation is bounded and the representers are $K(\cdot,t_j)\in\Hcal$, i.e., $S_m:\mathcal{H}_K\!\to\!\R^m$ can be defined as
\begin{equation}
(S_mv)_j \;=\; v(t_j) \;=\; \ip{v}{K(\cdot,t_j)},\qquad j=1,\dots,m,
\end{equation}
so the adjoint is $S_m^* a \;=\; \sum_{j=1}^m a_j\,K(\cdot,t_j)$. A convenient class of admissible nonlinearities is given by \cite{parada2025convolutionalrkhs}
\begin{equation}
\sigma(v) \;=\; S_m^*\,\sigma_\ast(S_mv)
\;=\; \sum_{j=1}^m \sigma_\ast\!\big((S_mv)_j\big)\,K(\cdot,t_j),
\end{equation}
with $\sigma_\ast:\R\to\R$ Lipschitz.

\end{document}